\documentclass{article} 
\usepackage{iclr2027_conference,times}

\usepackage{amsmath,amsfonts,bm}

\def\eqref#1{equation~\ref{#1}}

\def\1{\bm{1}}

\def\mA{{\bm{A}}}
\def\mB{{\bm{B}}}
\def\mC{{\bm{C}}}

\def\mF{{\bm{F}}}
\def\mG{{\bm{G}}}
\def\mH{{\bm{H}}}
\def\mI{{\bm{I}}}

\def\mL{{\bm{L}}}
\def\mM{{\bm{M}}}

\def\mP{{\bm{P}}}
\def\mQ{{\bm{Q}}}
\def\mR{{\bm{R}}}
\def\mS{{\bm{S}}}

\def\mU{{\bm{U}}}
\def\mV{{\bm{V}}}
\def\mW{{\bm{W}}}
\def\mX{{\bm{X}}}
\def\mY{{\bm{Y}}}
\def\mZ{{\bm{Z}}}

\DeclareMathAlphabet{\mathsfit}{\encodingdefault}{\sfdefault}{m}{sl}
\SetMathAlphabet{\mathsfit}{bold}{\encodingdefault}{\sfdefault}{bx}{n}

\usepackage{hyperref}
\usepackage{url}
\usepackage{bm}
\usepackage{booktabs}
\usepackage{amsmath, amssymb}
\usepackage{graphicx}
\usepackage{subcaption}
\usepackage{wrapfig}
\usepackage[linesnumbered,ruled,vlined]{algorithm2e}
\makeatletter
\renewcommand{\BlankLine}{\vskip 0.3ex}
\makeatother

\title{Rotated Manifold Optimization for Low-Rank Adaptation}

\author{Yuhui Ding\thanks{Work done during Yuhui's internship at Microsoft Research Cambridge.} \\ ETH Zurich \\ \texttt{yuhui.ding@inf.ethz.ch}
\And Javier Zazo \\ Microsoft Research Cambridge \\ \texttt{javierzazo@microsoft.com}
\And James Hensman \\ Microsoft Research Cambridge \\ \texttt{jameshensman@microsoft.com}
}

\newcommand{\grad}{\text{grad}}
\newcommand{\proj}{\text{Proj}}
\newcommand{\diag}{\text{diag}}
\newcommand{\rank}{\text{rank}}
\renewcommand{\eqref}[1]{Eq.~\ref{#1}}
\newcommand{\std}[1]{{\scriptsize$\pm$#1}}
\newcommand{\model}{RoM~}

\iclrfinalcopy 
\begin{document}

\maketitle

\begin{abstract}
We propose a novel optimizer for low-rank adaptation (LoRA) that explicitly incorporates the gauge symmetry of low-rank factorization. Our optimizer extends recent matrix optimizers for full-parameter training to the manifold of fixed-rank matrices by interpreting them as normalization under a rotated basis. We show how rotation and normalization can be integrated with the fixed-rank manifold efficiently. Our optimizer converges faster to lower held-out loss and achieves better or comparable downstream performance on both supervised finetuning and reinforcement learning tasks.
\end{abstract}

\section{Introduction}
\label{sec:intro}
Large language models (LLMs) based on the transformer architecture~\citep{vaswani2017attention} have achieved remarkable success in recent years~\citep{brown2020language, touvron2023llama, guo2025deepseek}. 
These models typically adopt a pretraining-finetuning paradigm, 
where large-scale pretraining equips a model with general capabilities and subsequent finetuning adapts it to specific downstream applications. However, as LLMs grow to billions of parameters, full-parameter finetuning becomes expensive in both computation and memory. Such a challenge has motivated the development of parameter-efficient finetuning (PEFT) methods~\citep{houlsby2019parameter}, which only update a small fraction of all parameters. Among these approaches, low-rank adaptation (LoRA)~\citep{hu2022lora} has emerged as a particularly effective method and has been widely adopted. 
LoRA freezes the pretrained weights and parametrizes the weight updates as products of low-rank matrices, substantially reducing the number of trainable parameters and the memory footprint, while maintaining competitive performance.

Compared with full-parameter training, LoRA exhibits a unique geometric structure induced by its low-rank factorization. Specifically, the low-rank matrices have infinitely many reparameterizations that represent the same weight update, 
since the model's output depends on these low-rank factors only through their product. 
This invariance, commonly referred to as gauge symmetry, suggests that optimization should respect the underlying geometry of the weight space of the LoRA parametrization. Motivated by this observation, we revisit the optimization of LoRA and aim to explicitly incorporate its gauge symmetry into the design of the optimization algorithm.

We design our optimizer on the manifold of fixed-rank matrices, aiming to optimize the product of the low-rank factors directly rather than each factor separately. We extend the recent advances in matrix optimizers~\citep{gong2026aro, jordan2024muon, vyas2025soap} to the fixed-rank manifold by interpreting their update rules as normalization under a rotated basis. 
Our optimizer applies row and column normalization in a conditioning frame defined by the left singular vectors of the tangent momentum, projects the resulting direction back onto the tangent space, and retracts to the fixed-rank manifold. 
By exploiting the low-rank structure of tangent vectors, all operations can be performed using thin factors without materializing full weight-sized matrices. 
We refer to our proposed optimizer as \model (\underline{Ro}tated \underline{M}anifold Optimization). 
We evaluate \model on supervised finetuning and reinforcement learning tasks. \model achieves substantially lower held-out loss and better or comparable downstream task performance compared with other LoRA optimizers, with only modest additional computational overhead.

\section{Background}
We briefly review the rotated optimization framework~\citep{gong2026aro} and LoRA~\citep{hu2022lora}, and then introduce the basic concepts of Riemannian optimization. Notations defined in this section will be used throughout the paper. 
We discuss related work in Appendix~\ref{sec:related_work}.

\subsection{Rotated optimization}
\label{sec:aro}
Recent matrix optimizers~\citep{jordan2024muon, vyas2025soap} can be viewed through a rotated optimization framework~\citep{gong2026aro}. Given a matrix-valued momentum $\mM \in \mathbb{R}^{m \times n}$, the framework first expresses $\mM$ in a rotated coordinate system defined by a rotation matrix $\mR \in \text{SO}(m)$, applies a normalization function $h$ in the rotated coordinates, and then rotates the result back to the original coordinates. The resulting update of the weight matrix $\mW$ takes the form:
\begin{equation}
\label{eq:aro}
    \Delta \mW \propto - \mR h\big(\mR^{\top}\mM\big)\,.
\end{equation}

\subsection{Low-rank adaptation (LoRA)}
\label{sec:lora}
LoRA~\citep{hu2022lora} freezes the pretrained weight matrix $\mW_0 \in \mathbb{R}^{m \times n}$ and parametrizes the weight adaptation $\mW_{\text{adp}}$ as the product of two low-rank factors:
\begin{equation}
\label{eq:lora}
    \bm{W}_{\text{eff}} = \mW_0 + \mW_{\text{adp}} = \bm{W}_0 + \bm{A}\bm{B}^{\top}, ~~\bm{A} \in \mathbb{R}^{m \times r}, ~~\bm{B} \in \mathbb{R}^{n \times r}
\end{equation}
with $r \ll \min (m, n)$. $\bm{W}_0$ denotes the pretrained weight matrix and is kept frozen throughout LoRA finetuning. The loss function $\ell$ can thus be written as $\ell(\bm{A}, \bm{B})$.

Note that, given the pretrained weight $\bm{W}_0$, the output of the model depends only on the product of $\bm{A}$ and $\bm{B}^{\top}$. Therefore, there exist many reparametrizations of $\bm{A}$ and $\bm{B}$ that keep $\ell$ unchanged:
\begin{equation}
\label{eq:gauge}
    \ell(\bm{A}, \bm{B}) = \ell(\bm{A}\bm{Q}^{-1}, \bm{B}\bm{Q}^{\top}), \quad \forall~\bm{Q} \in \text{GL}(r)
\end{equation}
where $\text{GL}(r)$ denotes the set of all $r \times r$ invertible matrices. In this paper, we propose an optimizer that explicitly incorporates the gauge symmetry of the LoRA parametrization (Eq.~\ref{eq:gauge}) into its design.

\subsection{Riemannian optimization}
\label{sec:rie_opt}
Let $\mathcal{M}$ be a smooth manifold~\citep{boumal2023intromanifolds}. We denote its tangent space at $\bm{w} \in \mathcal{M}$ by $\mathrm{T}_{\bm{w}} \mathcal{M}$. 
Intuitively, $\mathrm{T}_{\bm{w}} \mathcal{M}$ is a linear approximation of $\mathcal{M}$ locally at $\bm{w}$, and consists of all the velocities of smooth curves on $\mathcal{M}$ passing through $\bm{w}$. We use $\mathrm{T}\mathcal{M}$ to denote the tangent bundle, i.e., $\mathrm{T}\mathcal{M} = \{(\bm{w}, \bm{v})~\vert~ \bm{w} \in \mathcal{M}, \bm{v} \in \mathrm{T}_{\bm{w}}\mathcal{M}\}$. A retraction $\Phi: \mathrm{T}\mathcal{M} \rightarrow \mathcal{M}, ~(\bm{w}, \bm{v}) \mapsto \Phi_{\bm{w}}(\bm{v})$ is a smooth map such that it defines a smooth curve $t \mapsto \Phi_{\bm{w}}(t\bm{v})$ on $\mathcal{M}$ that satisfies $\Phi_{\bm{w}}(\bm{0}) = \bm{w}$ and $\frac{d}{dt}\Phi_{\bm{w}}(t\bm{v}) \vert_{t=0} = \bm{v}$.

We consider minimizing an objective $f(\bm{w})$ over $\mathcal{M}$. Since $\mathcal{M}$ is generally not a vector space, the notion of a gradient depends on how we measure the lengths and angles of tangent vectors. We therefore equip each tangent space $\mathrm{T}_{\bm{w}}\mathcal{M}$ with an inner product $\langle \bm{u}, \bm{v} \rangle_{\bm{w}}$. Intuitively, this inner product specifies the local geometry of $\mathcal{M}$ around $\bm{w}$ (i.e., how to compute local norms and angles). If the assignment $\bm{w} \mapsto \langle \cdot, \cdot \rangle_{\bm{w}}$ varies smoothly with $\bm{w}$, it is called a Riemannian metric, and $\mathcal{M}$ is referred to as a Riemannian manifold.

The Riemannian gradient of $f$ at $\bm{w}$, denoted by $\grad f(\bm{w})$, is defined as:
\begin{equation}
\label{eq:define_rie_gradient}
    \langle \grad f(\bm{w}), \bm{v}\rangle_{\bm{w}} = 
    \frac{d}{dt} f(c(t)) \Big\vert_{t=0}
    , \quad \forall~\bm{v} \in \mathrm{T}_{\bm{w}}\mathcal{M}
\end{equation}
where $c: \mathbb{R} \rightarrow \mathcal{M}$ is any smooth curve on $\mathcal{M}$ satisfying $c(0) = \bm{w}$ and $\frac{d}{dt}c(t) \vert_{t=0} = \bm{v}$. 
In practice, it is convenient to work with an extension $\Bar{f}$ defined in an open neighborhood $\Bar{\mathcal{M}}$ of $\mathcal{M}$ such that $f$ is the restriction of $\Bar{f}$ to $\mathcal{M}$, i.e., $f = \Bar{f}\vert_{\mathcal{M}}$. Let $\nabla \Bar{f}(\bm{w})$ denote the Euclidean gradient of $\Bar{f}$ at $\bm{w}$, computed in the ambient space (e.g., via automatic differentiation in PyTorch). When $\mathcal{M}$ is an embedded submanifold of $\mathbb{R}^n$ and inherits the Euclidean inner product, i.e., $\langle \bm{u}, \bm{v} \rangle_{\bm{w}} = \bm{u}^{\top}\bm{v}$, we have the following relationship between $\grad f(\bm{w})$ and $\nabla \Bar{f}(\bm{w})$:
\begin{equation}
\label{eq:define_proj}
    \grad f(\bm{w}) = \proj_{\bm{w}}\left(\nabla \Bar{f}(\bm{w})\right)
\end{equation} 
where $\proj_{\bm{w}}: \mathbb{R}^n \rightarrow \mathrm{T}_{\bm{w}}\mathcal{M}$ denotes the orthogonal projection onto $\mathrm{T}_{\bm{w}}\mathcal{M}$. Finally, given a step size $\eta>0$, a basic Riemannian gradient descent update takes the form:
\begin{equation}
    \bm{w}_{k+1} = \Phi_{\bm{w}_k}\big(-\eta\,\grad f(\bm{w}_k)\big)\,.
\end{equation}
\section{Method}
\label{sec:method}
We propose Rotated Manifold Optimization (RoM),
which directly optimizes the LoRA weight adaptation $\mW_{\text{adp}}$ on the manifold of fixed-rank matrices. 
For notational simplicity, throughout this section we write $\mW := \mW_{\text{adp}}$. 
We first introduce the geometry of the fixed-rank manifold in Section~\ref{sec:fixed-rank} and describe how to perform Riemannian optimization efficiently by exploiting the low-rank structure. Then in Section~\ref{sec:rot_opt} we extend the rotated optimization framework (\eqref{eq:aro}) to the manifold.

\subsection{Fixed-rank manifold}
\label{sec:fixed-rank}
We consider the set of all $m \times n$ matrices with rank $r$:
\begin{equation}
    \mathcal{M}_r = \{\bm{W} \in \mathbb{R}^{m \times n} ~\vert~ \text{rank}(\bm{W}) = r\}
\end{equation}
Each matrix $\bm{W} \in \mathcal{M}_r$ admits an SVD decomposition $\bm{W} = \bm{U}\bm{\Sigma}\bm{V}^{\top}$,
where $\bm{U} \in \mathbb{R}^{m \times r}$ and $\bm{V} \in \mathbb{R}^{n \times r}$ have orthonormal columns, i.e., $\bm{U}^{\top}\bm{U} = \bm{V}^{\top}\bm{V} = \bm{I}_r$, and $\bm{\Sigma} = \diag(\bm{\sigma})$ is a square diagonal matrix with positive diagonal entries $\bm{\sigma} \in \mathbb{R}_+^r$.
We parametrize $\mW$ using $\bm{U}$, $\bm{\sigma}$ and $\mV$.
Compared with the $\bm{A}$ and $\bm{B}$ factors in standard LoRA (\eqref{eq:lora}), we only need negligible additional memory to store $\bm{\sigma}$.

\paragraph{Tangent space}
Given the factors $\bm{U}$ and $\bm{V}$, the tangent space at $\bm{W}$ takes the form:
\begin{equation}
\label{eq:fr_tangent_space}
\begin{aligned}
\mathrm{T}_{\mW}\mathcal{M}_r
= \Big\{ \bm{Z} \in \mathbb{R}^{m \times n} ~\Big|~
&\bm{Z} = \mU \mS \mV^{\top} + \bm{U}_p\mV^{\top} + \mU \bm{V}_p^{\top}, \\
&\mS \in \mathbb{R}^{r \times r},~ \bm{U}_p \in \mathbb{R}^{m \times r},~ \bm{V}_p \in \mathbb{R}^{n \times r}, \\
&\mU^{\top}\bm{U}_p = \bm{0},~ \mV^{\top}\bm{V}_p = \bm{0}
\Big\}.
\end{aligned}
\end{equation}
\eqref{eq:fr_tangent_space} implies that, although a tangent vector $\bm{Z} \in \mathrm{T}_{\mW}\mathcal{M}_r$ has shape $m \times n$ in the ambient space, it can actually be represented and stored using smaller components $\mS \in \mathbb{R}^{r \times r}$, $\mU_p \in \mathbb{R}^{m \times r}$ and $\mV_p \in \mathbb{R}^{n \times r}$.
A tangent vector $\mZ$ defined as in \eqref{eq:fr_tangent_space} can be further written in the compact form:
\begin{equation}
\label{eq:tangent_matrix_form}
    \bm{Z} = \begin{bmatrix}
        \mU & \mU_p
    \end{bmatrix} \begin{bmatrix}
        \mS & \mI_r \\ \mI_r & \bm{0} 
    \end{bmatrix} \begin{bmatrix}
        \mV & \mV_p
    \end{bmatrix}^{\top}
\end{equation}
which indicates $\rank (\mZ) \leq 2r$ for any $\mZ \in \mathrm{T}_{\mW}\mathcal{M}_r$.

\paragraph{Orthogonal projection}
Let $\mathcal{L}: \mathcal{M}_r \rightarrow \mathbb{R}$ denote the objective to minimize and $\Bar{\mathcal{L}}$ a smooth extension of $\mathcal{L}$ to an open neighborhood of $\mathcal{M}_r$ in the ambient space.
Consistent with Section~\ref{sec:rie_opt}, we denote the Euclidean gradient and the Riemannian gradient at $\mW$ by  $\nabla\Bar{\mathcal{L}}(\mW)$ and $\grad\,\mathcal{L}(\mW)$ respectively.
We equip $\mathcal{M}_r$ with the Riemannian metric induced by the Euclidean inner product. 
Then we derive the orthogonal projection of $\nabla \Bar{\mathcal{L}}(\mW) \in \mathbb{R}^{m \times n}$ onto the tangent space $\mathrm{T}_{\mW}\mathcal{M}_r$:
\begin{equation}
\label{eq:fr_projection}
\proj_{\mW}\big(\nabla \Bar{\mathcal{L}}(\mW)\big)
= \nabla \Bar{\mathcal{L}}(\mW)
- (\mI - \mU\mU^{\top})\,\nabla \Bar{\mathcal{L}}(\mW)\,(\mI - \mV\mV^{\top})\,,
\end{equation}
and we have $\grad\,\mathcal{L}(\mW) = \proj_{\mW}\big(\nabla \Bar{\mathcal{L}}(\mW)\big)$.
Because $\proj_{\mW}\big(\nabla \Bar{\mathcal{L}}(\mW)\big) \in \mathrm{T}_{\mW}\mathcal{M}_r$, it admits the decomposition in \eqref{eq:fr_tangent_space}. Its three components can be expressed in terms of $\nabla \Bar{\mathcal{L}}(\mW), \mU, \mV$ as:
\begin{equation}
\label{eq:fr_projection_components}
\begin{aligned}
    \mS   &= \mU^{\top}\,\nabla \Bar{\mathcal{L}}(\mW)\,\mV\,, \\
    \mU_p &= \nabla \Bar{\mathcal{L}}(\mW)\,\mV - \mU\mS\,, \\
    \mV_p &= \nabla \Bar{\mathcal{L}}(\mW)^{\top}\,\mU - \mV\mS^{\top}\,.
\end{aligned}
\end{equation}

\paragraph{Practical Riemannian gradient computation}
While the relationship between the Euclidean gradient and the Riemannian gradient at $\mW$ is straightforward (\eqref{eq:define_proj}), naively applying the projection \eqref{eq:fr_projection} to the Euclidean gradient $\nabla \Bar{\mathcal{L}}(\mW)$ would require materializing a full $m \times n$ matrix, which is memory intensive. 
We demonstrate how to compute the Riemannian gradient efficiently via a lightweight modification to the auto-differentiation, without instantiating $\nabla \Bar{\mathcal{L}}(\mW)$ explicitly.

Consider $\mW \in \mathbb{R}^{m \times n}$ as the weight adaptation of a linear layer that takes $\mX \in \mathbb{R}^{N \times m}$ as input and outputs $\mY \in \mathbb{R}^{N \times n}$, where $N$ represents the effective batch size (i.e., the total number of tokens in the mini-batch).
In the standard auto-differentiation framework,
$\mX$ is saved during the forward pass for gradient computation, and when the gradient with respect to $\mY$, denoted by $\mG_Y = \partial \Bar{\mathcal{L}} / \partial \mY \in \mathbb{R}^{N \times n}$ is back-propagated to this layer, $\nabla \Bar{\mathcal{L}}(\mW)$ is computed as 
$\nabla \Bar{\mathcal{L}}(\mW) = \mX^{\top}\mG_Y$.
Note that to compute $\grad\, \mathcal{L}(\mW)$,
we only need $\mS$, $\mU_p$ and $\mV_p$ in \eqref{eq:fr_projection_components},
which depend on $\nabla \Bar{\mathcal{L}}(\mW)\mV$ and $\nabla \Bar{\mathcal{L}}(\mW)^{\top}\mU$ rather than any standalone $\nabla \Bar{\mathcal{L}}(\mW)$.
We can compute these two quantities directly without the need to materialize the $m \times n$ matrix 
$\nabla \Bar{\mathcal{L}}(\mW)$:
\begin{equation}
\label{eq:efficient_rie_opt}
    \nabla \Bar{\mathcal{L}}(\mW)\mV = \mX^{\top}\Big(\mG_Y\mV\Big) \in \mathbb{R}^{m \times r}
    \qquad 
    \nabla \Bar{\mathcal{L}}(\mW)^{\top}\mU = 
    \mG_Y^{\top}\Big(\mX\mU\Big) \in \mathbb{R}^{n \times r}
\end{equation}
We compute the bracketed products first, so that every matrix multiplication involves an operand with one dimension equal to $r$, keeping all intermediate matrices low-rank and memory-efficient.
We customize the backward propagation (e.g., via PyTorch's backward hooks) to compute $\nabla \Bar{\mathcal{L}}(\mW)\mV$ and $\nabla \Bar{\mathcal{L}}(\mW)^{\top}\mU$ on the fly as in \eqref{eq:efficient_rie_opt}, and store them for retraction. 
The memory usage is of the same order as that for computing and storing the gradients of $\mA$ and $\mB$ in standard LoRA (\eqref{eq:lora}).

\paragraph{Retraction}
Given a tangent update $\mZ \in \mathrm{T}_{\mW}\mathcal{M}_r$, 
we use the rank-$r$ truncated SVD as the retraction, i.e., $\Phi_{\mW}(\mZ) = \text{SVD}_r(\mW + \mZ)$, which retains the top $r$ singular values of $\mW + \mZ$ and their corresponding singular vectors. 
With the factorization of $\mZ$ in \eqref{eq:tangent_matrix_form},
$\mW + \mZ$ can be written in a compact form that enables efficient computation of the truncated SVD:
\begin{equation}
\label{eq:retraction_matrix_form}
    \mW + \mZ = \begin{bmatrix}
        \mU & \mU_p
    \end{bmatrix} \begin{bmatrix}
        \bm{\Sigma} + \mS & \mI_r \\ \mI_r & \bm{0} 
    \end{bmatrix} \begin{bmatrix}
        \mV & \mV_p
    \end{bmatrix}^{\top}\,.
\end{equation}
We apply thin QR factorizations $\mU_p = \mQ_U\mR_U$ and $\mV_p = \mQ_V\mR_V$. 
Then the truncated SVD is only applied to the $2r \times 2r$ matrix $\begin{bmatrix}
    \bm{\Sigma} + \mS & \mR_V^{\top} \\ \mR_U & \bm{0} 
\end{bmatrix}$. 
Let $\widetilde{\mU}\widetilde{\bm\Sigma}\widetilde{\mV}^{\top}$ denote the rank-$r$ truncated SVD of this small matrix, then the retracted point can be written as:
\begin{equation}
    \Phi_{\mW}(\mZ) = ([\mU~~\mQ_U]\widetilde{\mU})\widetilde{\bm{\Sigma}}
    ([\mV~~\mQ_V]\widetilde{\mV})^{\top}
\end{equation}
which gives a best rank-$r$ approximation of $\mW + \mZ$ in Frobenius norm.

\paragraph{Momentum transport}
A natural extension of the standard (Euclidean) momentum to the manifold suggests taking the moving average of the Riemannian gradients. However, Riemannian gradients of an optimization trajectory live in different tangent spaces and cannot be combined directly. 
Let $\mM_k \in \mathrm{T}_{\mW_k}\mathcal{M}_r$ be the momentum at the $k$-th step of an optimization trajectory, and let $\grad\,\mathcal{L}(\mW_{k+1})$ be the fresh Riemannian gradient at $\mW_{k+1}$. To get the updated momentum $\mM_{k+1}$, we first project $\mM_k$ onto the new tangent space $\mathrm{T}_{\mW_{k+1}}\mathcal{M}_r$ and then combine the projection of $\mM_k$ with $\grad\,\mathcal{L}(\mW_{k+1})$:
\begin{equation}
\label{eq:momentum_transport}
\mM_{k \rightarrow k+1} = \proj_{\mW_{k+1}}\big(\mM_k\big)
\qquad
\mM_{k+1} = \beta \mM_{k \rightarrow k+1} + (1 - \beta)\grad\,\mathcal{L}(\mW_{k+1})
\end{equation}
Similar to the Riemannian gradient, the momentum is also stored using its factorization (\eqref{eq:fr_tangent_space}), and the transport operation (\eqref{eq:momentum_transport}) remains factorized. A detailed algorithm on \eqref{eq:momentum_transport} is provided in Algorithm~\ref{alg:fixed-rank-momentum} in the appendix.

\subsection{Rotated optimization on manifold}
\label{sec:rot_opt}

\begin{algorithm}[t]
\caption{Rotated optimization on the fixed-rank manifold}
\label{alg:rotate-precondition-project}
\DontPrintSemicolon
\KwIn{Weight adaptation $\mW = \mU\bm{\Sigma}\mV^{\top}$; 
momentum $\mM\in \mathrm{T}_{\mW}\mathcal{M}_r$ represented by components $(\mS,\mU_p,\mV_p)$; conditioning frame $\mF\in\mathbb{R}^{m\times 2r}$ with $\mF^\top\mF=\bm{I}_{2r}$; normalization function $h$.}
\KwOut{Components of the final tangent update: $(\widetilde{\mS},\widetilde{\mU}_p,\widetilde{\mV}_p)$}
\BlankLine
\tcp*[l]{Form thin factors $\mL,\mR$ of $\mM$ from its components ($\mM=\mL\mR^\top$)}
$\mL \leftarrow \begin{bmatrix} \mU\mS+\mU_p & \mU \end{bmatrix} \in \mathbb{R}^{m \times 2r}$\;
$\mR \leftarrow \begin{bmatrix} \mV & \mV_p \end{bmatrix} \in \mathbb{R}^{n \times 2r}$\;
\BlankLine
\tcp*[l]{Express $\mM$ in the rotated coordinates: $\bm{C}=\mF^\top\mM$}
\tcp*[l]{Note the order of products that avoids $m \times n$ matrix}
$\bm{C} \leftarrow \left(\mF^\top\mL\right)\mR^\top \in \mathbb{R}^{2r \times n}$\;
\BlankLine
\tcp*[l]{Apply the normalization function $h$}
$\widetilde{\bm{C}} \leftarrow h(\bm{C})$\;
\BlankLine
\tcp*[l]{Map back to the original coordinates $\mF\widetilde{\bm{C}}$ and project}
\tcp*[l]{Matrix multiplications are reordered to avoid $m \times n$ matrix}
$\mP \leftarrow \mF^\top\mU \in \mathbb{R}^{2r \times r}$\;
$\mQ \leftarrow \widetilde{\bm{C}}\mV \in \mathbb{R}^{2r \times r}$\;
$\widetilde{\mS} \leftarrow \mP^\top\mQ$\;
$\widetilde{\mU}_p \leftarrow \mF\mQ-\mU\widetilde{\mS}$\;
$\widetilde{\mV}_p \leftarrow \widetilde{\bm{C}}^\top\mP-\mV\widetilde{\mS}^\top$\;
\end{algorithm}

Starting from the tangent momentum $\mM \in \mathrm{T}_{\mW}\mathcal{M}_r$ (\eqref{eq:momentum_transport}), we now extend the rotated optimization framework (\eqref{eq:aro}) to the manifold setting. This introduces two challenges. First, 
a naive rotation in the ambient space would require an $m \times m$ rotation matrix ($O(m^2)$ memory), destroying the memory efficiency of LoRA. 
Second, after rotation and normalization, the resulting vector may not remain in the tangent space. We show that both issues can be addressed efficiently by using the rank-$2r$ structure of tangent vectors.

\paragraph{Conditioning frame}
Since the momentum $\mM \in \mathrm{T}_{\mW}\mathcal{M}_r$ has rank at most $2r$, 
its column space is low-dimensional. Therefore, it suffices to perform the rotation within a $2r$-dimensional subspace containing the column space of $\mM$,
rather than in the full ambient space\footnote{Any rotation acting only within the orthogonal complement of $\mM$'s column space leaves $\mM$ unchanged.}. 
Concretely, we construct a conditioning frame $\mF \in \mathbb{R}^{m \times 2r}$ 
with orthonormal columns (i.e., $\mF^{\top}\mF = \mI_{2r}$) for this subspace. The frame $\mF$ defines the rotated coordinate system, in which $\mM$ is represented by $\mF^{\top}\mM$.
In full-parameter training, the left singular vectors of $\mM$ have been chosen by several previous works \citep{jordan2024muon, vyas2025soap} to define the rotated coordinate system, but computing them can be expensive.
When $\mM \in \mathrm{T}_{\mW}\mathcal{M}_r$, its low-rank structure (\eqref{eq:tangent_matrix_form}) allows us to compute these singular vectors efficiently using two thin QR factorizations and an SVD of a $2r \times 2r$ matrix, analogous to how we compute the retraction in Section~\ref{sec:fixed-rank}. We therefore use the resulting left singular vectors as the conditioning frame $\mF$.

\paragraph{Normalization function}
We next specify the normalization function $h$ applied to the momentum in the rotated coordinates. 
Note that the normalization is applied to the reduced representation $\mF^{\top}\mM \in \mathbb{R}^{2r \times n}$, rather than to a full 
$m \times n$ matrix. 
Following~\cite{scetbon2025gradient}, we adopt a multi-norm normalization that has shown effectiveness in LLM pre-training~\citep{gong2026aro}. 
For any matrix $\mG \in \mathbb{R}^{m \times n}$, 
$h$ is defined as:
\begin{equation}
\label{eq:sinkhorn}
h(\mG) = \operatorname*{arg\,max}_{\mH \in \mathbb{R}^{m \times n}}\,\langle \mG, \mH\rangle \quad \text{s.t.} \quad \max_i \|\mH_{i, :}\|_2 = 1 \quad \text{and} \quad \max_j \|\mH_{:, j}\|_2 = 1
\end{equation}
Here, $\mH_{i, :}$ and $\mH_{:,j}$ denote the $i$-th row ($1 \leq i \leq m$) and $j$-th column ($1 \leq j \leq n$) of $\mH$ respectively; 
$\langle \mG, \mH \rangle = \text{tr}(\mG \mH^{\top})$ denotes the Frobenius inner product; 
and $\|\cdot\|_2$ denotes the $\ell_2$ norm. 
\eqref{eq:sinkhorn} generalizes the single-norm steepest-descent formulation of \citet{bernstein2024old} to simultaneous row- and column-norm constraints. 
In practice, we approximate \eqref{eq:sinkhorn} by alternating row and column normalization for a small fixed number of iterations $K$, which we set to $5$ throughout all experiments. 
The full procedure is given in Algorithm~\ref{alg:sinkhorn} in the appendix.

\paragraph{Final update rule}
Given the conditioning frame $\mF$ and tangent momentum $\mM \in \mathrm{T}_{\mW}\mathcal{M}_r$, we first represent $\mM$ in the rotated coordinates as $\mF^{\top}\mM$, apply the normalization function $h$, and then 
map the resulting direction back to the original coordinates as $\mF h(\mF^{\top}\mM)$. 
Although $\mM$ lies in the tangent space $\mathrm{T}_{\mW}\mathcal{M}_r$, 
the normalization function $h$ does not, in general, preserve this tangent space constraint. 
We therefore project the resulting direction back onto $\mathrm{T}_{\mW}\mathcal{M}_r$. 
With a step size $\eta > 0$, 
the final tangent update $\bm{\xi} \in \mathrm{T}_{\mW}\mathcal{M}_r$ is:
\begin{equation}
\label{eq:final_update}
\bm{\xi} = -\,\eta\,\proj_{\mW}\Big(\mF h\Big(\mF^{\top}\mM\Big)\Big)
\end{equation}
Importantly, the orthogonal projection in \eqref{eq:final_update} can be combined with the preceding operations so that $\mF h(\mF^{\top}\mM)$ is never materialized as a full $m \times n$ matrix. We detail the efficient implementation of one step of \model in Algorithm~\ref{alg:rotate-precondition-project}.

\section{Experiments}
\label{sec:exp}
We evaluate \model on supervised finetuning (SFT) and reinforcement learning (RL) tasks, comparing its optimization and downstream performance against existing LoRA optimizers. We describe the experimental setup in Section~\ref{sec:exp_setup}, 
present the SFT and RL results in Section~\ref{sec:exp_sft} and~\ref{sec:exp_rl}, respectively, 
and further study the contribution of its components and its sensitivity to learning-rate selection in Section~\ref{sec:exp_ablation}.

\begin{figure}[t]
\centering
\includegraphics[width=0.804\textwidth]{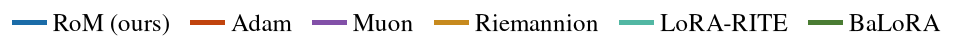}\\[2pt]
\begin{subfigure}[b]{0.45\textwidth}%
    \includegraphics[width=\linewidth]{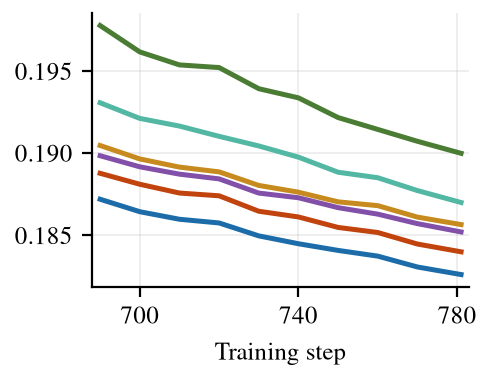}
    \caption{Llama-3.2-3B}
    \label{fig:valcurve_r16_llama3b}
\end{subfigure}%
\hspace{0.05\textwidth}%
\begin{subfigure}[b]{0.45\textwidth}%
    \includegraphics[width=\linewidth]{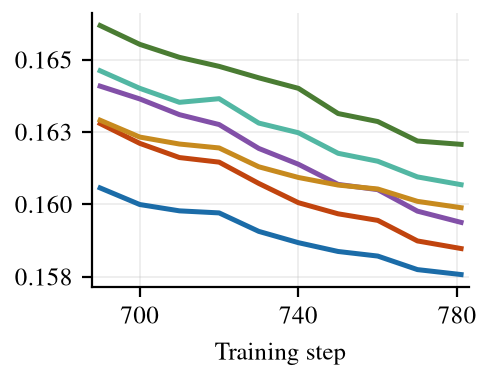}
    \caption{Llama-3.1-8B}
    \label{fig:valcurve_r16_llama8b}
\end{subfigure}%
\caption{Validation loss (mean cross-entropy across response tokens) over the last 100 training steps on the math SFT task at rank $r=16$ (see Appendix~\ref{sec:exp_appendix} for the $r=32$ case and additional results across different ranks).}
\label{fig:valcurve_r16}
\vspace{-4mm}
\end{figure}

\subsection{Experimental setup}
\label{sec:exp_setup}
We apply LoRA adapters $\mW = \mU\bm{\Sigma}\mV^{\top}$ to the weight matrices of all linear layers in \textit{both} the attention and MLP modules, while keeping one-dimensional parameters, input embeddings and the output head frozen. We initialize $\mU$ and $\mV$ randomly with orthonormal columns, and set $\bm{\Sigma} = 10^{-3}\mI_r$. This initialization ensures that each adapter is a valid point on the fixed-rank manifold, while keeping the effective weights close enough to their pretrained values.

We compare \model with Adam~\citep{kingma2014adam}, Muon~\citep{jordan2024muon}, 
LoRA-RITE~\citep{yen2025lora}, 
BaLoRA~\citep{castin2026balanced}, 
and Riemannion~\citep{bogachev2026lora}. 
Adam and Muon adopt the conventional two-factor LoRA (\eqref{eq:lora}) and optimize each factor separately. 
LoRA-RITE~\citep{yen2025lora} uses matrix preconditioners to achieve transformation invariance. 
BaLoRA~\citep{castin2026balanced} addresses the parametrization redundancy of LoRA (\eqref{eq:gauge}) through a balanced projection step after each Adam update to enforce $\mA^{\top}\mA = \mB^{\top}\mB$. Riemannion~\citep{bogachev2026lora} projects each Muon update onto the tangent space of the fixed-rank manifold.

For SFT, we use the same learning rate schedule for all optimizers: linear warm-up over the first $5\%$ of training steps, followed by linear decay to $10\%$ of the peak learning rate. 
We find this warm-up schedule particularly beneficial for Adam, improving its performance compared with previously reported results. 
For RL finetuning, we follow the setup of~\citet{liu2025understanding} and use a constant learning rate. 
For each optimizer, 
we tune its learning rate over the same logarithmically spaced grid and expand the grid when necessary to ensure that the best-performing value is bracketed on both sides.
We use a momentum coefficient of 0.9 for all optimizers. For Adam, we set the second-moment coefficient to $\beta_2 = 0.95$. We use no weight decay.

\begin{table}[t]
\centering
\caption{Math SFT results for the Llama models at rank $r=16$: final MetaMathQA validation loss
($\times10^{-3}$) and test accuracies (\%). Results are mean $\pm$ standard
deviation over three seeds; the pretrained models are evaluated once. Bold marks
the best entry and entries whose intervals overlap it; no entry is bold if all intervals overlap.}
\label{tab:main_results_r16}
\small
\setlength{\tabcolsep}{4pt}
\begin{tabular}{@{}l ccc @{\hspace{14pt}} ccc@{}}
\toprule
& \multicolumn{3}{c}{Llama-3.2-3B} & \multicolumn{3}{c}{Llama-3.1-8B} \\
\cmidrule(lr){2-4} \cmidrule(lr){5-7}
& MetaMathQA val $\downarrow$ & GSM8K $\uparrow$ & MATH $\uparrow$ & MetaMathQA val $\downarrow$ & GSM8K $\uparrow$ & MATH $\uparrow$\\
\midrule
Pretrained & 728.23 & 18.1 & 7.3 & 652.22 & 41.9 & 15.5 \\
\midrule
RoM (ours) & \textbf{182.69}\std{0.21} & \textbf{65.4}\std{0.1} & \textbf{18.2}\std{0.3} & \textbf{157.60}\std{0.06} & 77.7\std{1.0} & \textbf{29.2}\std{0.3} \\
Adam       & 184.11\std{0.14} & \textbf{64.9}\std{0.9} & \textbf{18.0}\std{0.1} & 158.70\std{0.26} & 77.2\std{0.6} & 28.4\std{0.4} \\
Muon       & 185.42\std{0.19} & 64.3\std{0.9} & \textbf{17.8}\std{0.4} & 159.45\std{0.13} & 77.3\std{0.9} & 28.0\std{0.5} \\
Riemannion & 185.79\std{0.22} & 64.0\std{0.7} & \textbf{17.7}\std{0.4} & 159.91\std{0.13} & 76.9\std{0.7} & \textbf{29.0}\std{0.6} \\
LoRA-RITE & 187.16\std{0.16} & 64.5\std{0.4} & 17.2\std{0.5} & 160.54\std{0.16} & 76.6\std{0.8} & 28.0\std{0.7} \\
BaLoRA     & 190.31\std{0.28} & 62.8\std{1.1} & 17.3\std{0.2} & 161.93\std{0.12} & 76.1\std{0.8} & 28.0\std{0.2} \\
\bottomrule
\end{tabular}
\vspace{-4mm}
\end{table}

\subsection{Supervised finetuning}
\label{sec:exp_sft}

\paragraph{Mathematical reasoning}
We finetune Llama-3.2-3B, Llama-3.1-8B~\citep{grattafiori2024llama}, and
Gemma-2-27B~\citep{gemma2024gemma2} at rank $r=16$ on 100K 
MetaMathQA examples~\citep{yu2024metamath} for one epoch, using an effective batch size of 128. 
We construct a held-out 10K-example validation set and measure validation loss
as the mean cross-entropy over response tokens. 
For each optimizer, we
independently select the learning rate that minimizes validation loss and
repeat the selected configuration over three seeds. 
Training for additional epochs did not further reduce the final validation loss. 
We evaluate downstream
performance using zero-shot accuracy on GSM8K (1319 examples) and MATH
(5000 examples) with greedy decoding and bfloat16 precision. 
Figure~\ref{fig:valcurve_r16} shows the validation-loss trajectories for the two Llama models. Tables~\ref{tab:main_results_r16} and \ref{tab:gemma_math} report the mean and standard deviation of the final validation loss and downstream accuracies for the Llama models and Gemma-2-27B, respectively.

\begin{wraptable}{r}{0.5\textwidth}
\vspace{-6pt}
\centering
\small
\setlength{\tabcolsep}{4pt}
\captionof{table}{Math SFT results for Gemma-2-27B at rank 16. Bold follows the convention in Table~\ref{tab:main_results_r16}.}
\label{tab:gemma_math}
\vspace{-2mm}
\begin{tabular}{@{}l ccc@{}}
\toprule
& Math SFT val $\downarrow$ & GSM8K $\uparrow$ & MATH $\uparrow$ \\
\midrule
RoM (ours) & \textbf{126.19}\std{0.13} & 84.2\std{0.8} & \textbf{43.4}\std{0.3} \\
Adam & 126.55\std{0.21} & 84.1\std{0.6} & 42.2\std{0.5} \\
Muon & 127.02\std{0.29} & 83.8\std{0.7} & 42.6\std{0.4} \\
Riemannion & 127.85\std{0.20} & 83.8\std{0.9} & 42.2\std{0.6} \\
LoRA-RITE & 129.71\std{0.13} & 83.7\std{0.5} & 42.7\std{0.2} \\
BaLoRA & 129.05\std{0.27} & 84.6\std{0.9} & \textbf{43.3}\std{0.1} \\
\bottomrule
\end{tabular}
\vspace{-4mm}
\end{wraptable}

At all three model sizes, \model achieves the lowest final validation loss consistently, followed by Adam among the baselines\footnote{
Our BaLoRA results use the same training protocol as the other optimizers in our comparison, including the learning-rate schedule and independent learning-rate tuning, and are therefore not directly comparable to those reported by~\citet{castin2026balanced}.
}. 
\model obtains the highest mean accuracy on both downstream benchmarks for the Llama models,
although the differences are small relative to variation across seeds. On Gemma-2-27B,
\model yields the highest mean accuracy on MATH, 
while the standard-deviation intervals on GSM8K overlap across all
optimizers. In contrast to these small downstream differences, the
validation-loss gaps are much larger than the variation across seeds, showing a clear
optimization advantage for \model.

\begin{figure}[t]
\centering
\includegraphics[width=0.314\textwidth]{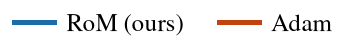}\\[2pt]
\hspace*{-0.123in}%
\begin{subfigure}[b]{0.333\textwidth}%
    \includegraphics[width=\linewidth]{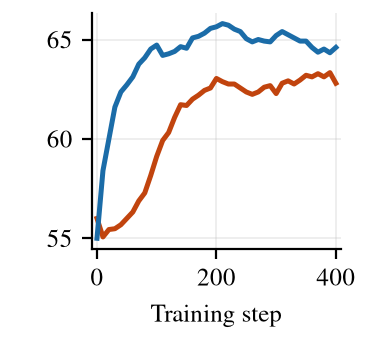}
    \caption{MATH-500}
    \label{fig:rl_math500}
\end{subfigure}%
\begin{subfigure}[b]{0.333\textwidth}%
    \includegraphics[width=\linewidth]{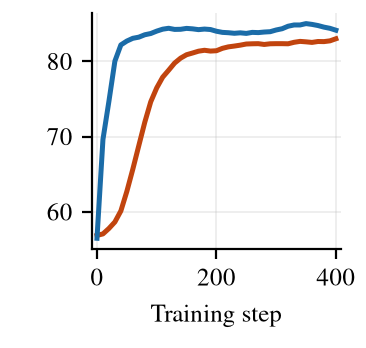}
    \caption{GSM8K}
    \label{fig:rl_gsm8k}
\end{subfigure}%
\begin{subfigure}[b]{0.333\textwidth}%
    \includegraphics[width=\linewidth]{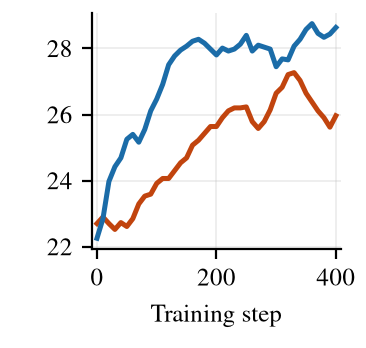}
    \caption{OlympiadBench}
    \label{fig:rl_olympiad}
\end{subfigure}
\caption{Evaluation accuracy (\%) during GRPO finetuning of Qwen2.5-3B with rank
$r=16$, on each of the three held-out benchmarks. 
The curves are smoothed using a 5-point centered moving average for visualization; higher accuracy is better.}
\label{fig:rl_curves}
\end{figure}

\begin{figure}[t]
\centering
\begin{minipage}[t]{0.45\textwidth}
    \vspace{0pt}
    \centering
    \includegraphics[width=\linewidth]{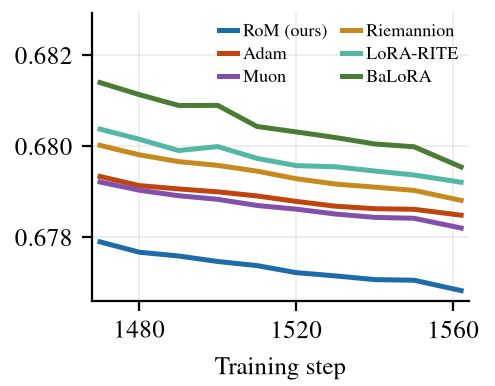}
    \caption{Validation loss over the final 100 training steps on the code SFT task, for Llama-3.1-8B at rank $r=32$.}
    \label{fig:valcurve_code}
\end{minipage}\hfill
\begin{minipage}[t]{0.53\textwidth}
    \vspace{2mm}
    \centering
    \small
    \setlength{\tabcolsep}{4pt}
    \captionof{table}{Code SFT results for Llama-3.1-8B at rank 32: final validation loss ($\times 10^{-3}$) on CodeFeedback and Pass@1 (\%) on HumanEval and MBPP. Results are mean $\pm$ standard deviation over 3 seeds except for the pretrained model. 
    Bold follows the convention in Table~\ref{tab:main_results_r16}.}
    \begin{tabular}{@{}lccc@{}}
    \toprule
    & Code SFT val $\downarrow$ & HumanEval $\uparrow$ & MBPP $\uparrow$\\
    \midrule
    Pretrained & 811.99 & 42.7 & 60.6 \\
    \midrule
    RoM (ours) & \textbf{676.81}\std{0.04} & 55.7\std{2.6} & 68.5\std{0.5} \\
    Adam       & 678.50\std{0.03} & 55.3\std{2.9} & 68.3\std{0.3} \\
    Muon       & 678.19\std{0.05} & 55.3\std{0.9} & 68.7\std{1.0} \\
    Riemannion & 678.87\std{0.06} & 55.7\std{1.4} & 68.2\std{0.4} \\
    LoRA-RITE  & 679.25\std{0.08} & 55.5\std{0.6} & 68.1\std{0.6} \\
    BaLoRA     & 679.63\std{0.15} & 56.7\std{1.0} & 68.6\std{1.0} \\
    \bottomrule
    \end{tabular}
    \label{tab:code_acc}
\end{minipage}
\vspace{-6mm}
\end{figure}

\paragraph{Code generation}
We finetune the Llama-3.1-8B base model on 100K examples of the CodeFeedback dataset~\citep{zheng2024opencodeinterpreter} using the Alpaca prompt templates~\citep{alpaca}. 
We compute validation loss as the mean cross-entropy over response tokens on a held-out validation set of 10K examples.
Since CodeFeedback sequences are typically longer than those in MetaMathQA, we train for one epoch with a smaller effective batch size of 64.
Figure~\ref{fig:valcurve_code} shows the validation loss over the final 100 training steps. \model achieves the lowest validation loss.

We evaluate the finetuned model on HumanEval~\citep{chen2021evaluating} (164 test examples) and MBPP~\citep{austin2021program} (378 test examples curated by EvalPlus~\citep{liu2023your}) using greedy decoding with bfloat16 precision and no in-context examples, and report Pass@1. 
For each optimizer, we train three runs with different random seeds. 
Table~\ref{tab:code_acc} reports the mean and standard deviation of the final validation loss and downstream performance. 
All optimizers clearly improve Pass@1 over the pretrained model. 
However, the differences in Pass@1 among optimizers are minor relative to the variation across random seeds, and we therefore do not observe a clear downstream advantage for any optimizer. 
In contrast, the validation loss gaps substantially exceed the seed variation, 
with \model achieving the lowest value. 
This suggests that RoM's optimization advantage extends to code SFT, although 
it does not translate into a clear Pass@1 improvement in this setting.

\subsection{Reinforcement learning}
\label{sec:exp_rl}

We further evaluate \model on LoRA-based RL finetuning using the VeRL framework~\citep{sheng2025hybridflow}. 
Since the other LoRA baselines do not specify an RL training recipe, we focus on comparing \model against Adam.
We train the Qwen2.5-3B base model~\citep{qwen2} using GRPO~\citep{shao2024deepseekmath} on the training split of MATH (7.5K problems) with rank $r=16$. 
At each rollout step, 
we sample 8 responses for each of 128 prompts, yielding 1024 sequences per optimization step.
We use no KL penalty and train for 400 steps. 
Following the setup of~\citet{liu2025understanding}, we use the bare questions as prompts with no chat template.
We evaluate the finetuned model using greedy decoding on three held-out benchmarks: MATH-500 (500 examples), GSM8K test (1319 examples), 
and OlympiadBench~\citep{he2024olympiadbench} (674 examples). 
We tune Adam's learning rate on a logarithmic grid and select $3\times 10^{-6}$. 
We use the same learning rate for \model without further tuning.

\begin{wraptable}{r}{0.5\textwidth}
\centering
\small
\setlength{\tabcolsep}{3pt}
\caption{Final downstream accuracy (\%) after GRPO finetuning over three independent runs. Results are mean $\pm$ standard deviation. ``Time'' denotes the median end-to-end wall clock time per optimizer step, normalized to Adam.}
\vspace{-4pt}
\label{tab:rl}
\begin{tabular}{@{}lcccc@{}}
\toprule
& MATH-500 & GSM8K & Olympiad & Time \\
\midrule
Pretrained  & 55.5 & 56.8 & 22.0 & -- \\
\midrule
Adam      & 62.9\std{0.3} & 82.7\std{0.4} & 25.9\std{0.8} & $1.00\times$ \\
RoM (ours) & \textbf{64.9}\std{0.6} & \textbf{84.2}\std{0.4} & \textbf{29.0}\std{0.8} & $1.08\times$ \\
\bottomrule
\end{tabular}
\end{wraptable}

Figure~\ref{fig:rl_curves} shows evaluation accuracy on the three held-out benchmarks throughout RL finetuning. 
\model improves accuracy faster than Adam and reaches higher final accuracy on all three benchmarks. 
Table~\ref{tab:rl} reports the final evaluation results over three independent runs. 
Unlike in SFT, 
where downstream differences among optimizers are relatively small, 
\model shows a clearer advantage over Adam in RL finetuning, 
while requiring only $1.08\times$ the per-step training time on 4 NVIDIA RTX PRO 6000 GPUs.

\subsection{Additional analysis}
\label{sec:exp_ablation}

\begin{figure}[t]
\centering
\begin{subfigure}[b]{0.45\textwidth}%
    \includegraphics[width=\linewidth]{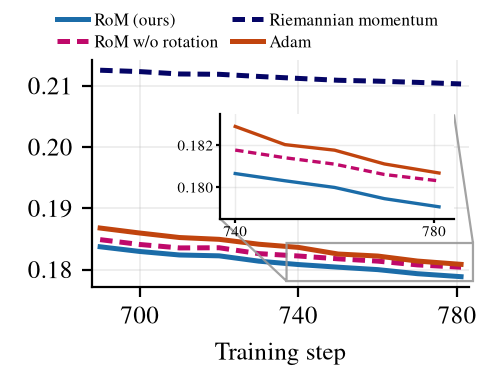}
    \caption{Ablation study}
    \label{fig:ablation}
\end{subfigure}%
\hspace{0.05\textwidth}%
\begin{subfigure}[b]{0.45\textwidth}%
    \includegraphics[width=\linewidth]{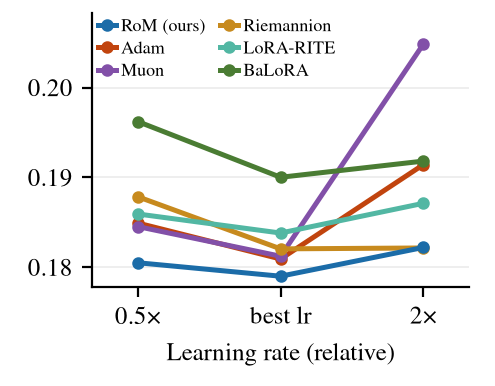}
    \caption{Learning rate sensitivity}
    \label{fig:lrsens_llama3b}
\end{subfigure}
\caption{Llama-3.2-3B at rank $r=32$ on the math SFT task.
(a) Validation loss over the last 100 training steps, removing parts of \model\!. (b) Validation loss at different ratios of best learning rate.}
\label{fig:ablation_lrsens}
\end{figure}

\paragraph{Ablation study}
We ablate the rotation and normalization components of \model on the math SFT task using Llama-3.2-3B at rank $r=32$.
We consider \model without rotation, which applies the normalization function (\eqref{eq:sinkhorn}) directly to the momentum (\eqref{eq:momentum_transport}), and \model without both rotation and normalization, which we refer to as \textsl{Riemannian momentum}.
We tune the learning rate for each variant following the same procedure as in Section~\ref{sec:exp_sft}. 
Figure~\ref{fig:ablation} shows the validation loss over the final 100 training steps. 
Riemannian momentum performs substantially worse than RoM, while adding normalization without rotation reduces the gap and brings its performance close to Adam. 
Full \model achieves the lowest validation loss, demonstrating the additional benefit of rotation.

\vspace{-0.5em}
\paragraph{Learning rate sensitivity}
We evaluate each optimizer at $0.5\times$, $1\times$, and $2\times$ its best learning rate under the same math SFT setting. 
Figure~\ref{fig:lrsens_llama3b} shows the final validation loss at each learning-rate ratio. 
The optimizers that account for the symmetry of LoRA's parametrization tend to exhibit less degradation away from their best learning rates than Adam and Muon. \model maintains the lowest validation loss across all three learning-rate ratios.

\section{Conclusion}
We presented RoM, a manifold-based optimizer for LoRA that directly optimizes the fixed-rank weight adaptation while avoiding the redundancy of factorized parameterizations. By exploiting the low-rank structure of tangent vectors, \model enables efficient Riemannian optimization with memory requirement comparable to conventional LoRA optimization. Across supervised finetuning and reinforcement learning experiments, \model consistently achieves lower validation loss than existing LoRA optimizers, while also improving downstream performance in the RL setting. Ablation experiments further show that both normalization and rotation contribute to its optimization performance.

\pagebreak
\subsection*{AI use statement}
During the preparation of this work, we used LLMs to assist with coding and debugging, 
to help document and organize experimental progress, 
and to polish the text and figures of the manuscript. 
The authors have reviewed and verified all LLM-assisted outputs and 
take full responsibility for the final content presented in this work.

\subsubsection*{Acknowledgments}
The authors thank Wenbo Gong and Chao Ma for their valuable feedback during the development of this work.

\bibliography{reference}
\bibliographystyle{iclr2027_conference}

\clearpage
\appendix

\section{Algorithms}
\begin{algorithm}[h]
\caption{Factorized momentum transport on the fixed-rank manifold}
\label{alg:fixed-rank-momentum}
\DontPrintSemicolon
\KwIn{Old and new orthonormal frames $\mU_k,\mV_k$, $\mU_{k+1},\mV_{k+1}$; momentum coefficient $\beta\in[0,1)$; old momentum components $(\mS_k,\mU_{p,k},\mV_{p,k})$ such that
$\mM_k = \mU_k\mS_k\mV_k^\top + \mU_{p,k}\mV_k^\top + \mU_k\mV_{p,k}^\top$, with $\mU_k^\top \mU_{p,k}=\bm{0}$ and $\mV_k^\top \mV_{p,k}=\bm{0}$; fresh Riemannian gradient components $(\mS^{(g)}_{k+1},\mU^{(g)}_{p,k+1},\mV^{(g)}_{p,k+1})$ at $\mW_{k+1}$.}
\KwOut{Updated momentum components $(\mS_{k+1},\mU_{p,k+1},\mV_{p,k+1})$ representing $\mM_{k+1}=\beta\,\proj_{\mW_{k+1}}(\mM_k) + (1-\beta)\grad\,\mathcal{L}(\mW_{k+1})$.}
\BlankLine
\tcp*[l]{Compute frame overlaps}
$\mC_U \leftarrow \mU_k^\top \mU_{k+1}$, $\mC_V \leftarrow \mV_k^\top \mV_{k+1}$\;
\BlankLine
\tcp*[l]{Compute $\bm{\Psi}_V=\mM_k\mV_{k+1}$ (no $m\times n$ matrix)}
$\bm{\Psi}_V \leftarrow \mU_k\big(\mS_k \mC_V + \mV_{p,k}^\top \mV_{k+1}\big) + \mU_{p,k} \mC_V$\;
\BlankLine
\tcp*[l]{Compute $\bm{\Psi}_U=\mM_k^\top\mU_{k+1}$ (no $m\times n$ matrix)}
$\bm{\Psi}_U \leftarrow \mV_k\big(\mS_k^\top \mC_U + \mU_{p,k}^\top \mU_{k+1}\big) + \mV_{p,k} \mC_U$\;
\BlankLine
\tcp*[l]{Project onto new tangent space}
$\mS_{k\rightarrow k+1} \leftarrow \mU_{k+1}^\top \bm{\Psi}_V$\;
$\mU_{p,k\rightarrow k+1} \leftarrow \bm{\Psi}_V - \mU_{k+1}\mS_{k\rightarrow k+1}$\;
$\mV_{p,k\rightarrow k+1} \leftarrow \bm{\Psi}_U - \mV_{k+1}\mS_{k\rightarrow k+1}^\top$\;
\BlankLine
\tcp*[l]{Accumulate with fresh Riemannian gradient}
$\mS_{k+1} \leftarrow \beta\,\mS_{k\rightarrow k+1} + (1-\beta)\,\mS^{(g)}_{k+1}$\;
$\mU_{p,k+1} \leftarrow \beta\,\mU_{p,k\rightarrow k+1} + (1-\beta)\,\mU^{(g)}_{p,k+1}$\;
$\mV_{p,k+1} \leftarrow \beta\,\mV_{p,k\rightarrow k+1} + (1-\beta)\,\mV^{(g)}_{p,k+1}$\;
\end{algorithm}

\begin{algorithm}[h]
\caption{Row and column normalization function $h$}
\label{alg:sinkhorn}
\DontPrintSemicolon
\KwIn{Reduced representation $\bm{C}\in\mathbb{R}^{2r\times n}$; number of iterations $K$.}
\KwOut{Normalized representation $\widetilde{\bm{C}}\in\mathbb{R}^{2r\times n}$.}
\BlankLine
\For{$k=1,\dots,K$}{
  $\bm{C}_{i,:} \leftarrow \bm{C}_{i,:} \,/\, \lVert \bm{C}_{i,:}\rVert_2$ \quad for $i=1,\dots,2r$\;
  $\bm{C}_{:,j} \leftarrow \bm{C}_{:,j} \,/\, \lVert \bm{C}_{:,j}\rVert_2$ \quad for $j=1,\dots,n$\;
}
\KwRet $\widetilde{\bm{C}} \leftarrow \bm{C}$\;
\end{algorithm}

\section{Related work}
\label{sec:related_work}
We review the relevant literature on optimization methods for LoRA. 
\citet{hayou2024lora} advocate setting different learning rates for the two low-rank factors to achieve efficient feature learning. 
\citet{zhang2024riemannian} extend this approach by introducing matrix preconditioners to stabilize feature learning and draw a connection to a Riemannian metric on a quotient manifold. 
\citet{yen2025lora} design preconditioners that make the update to the weight adapter (i.e., $\mW_{\text{adp}}$) invariant to reparametrizations of low-rank factors. 
\citet{castin2026balanced} enforce the condition 
$\mA^{\top}\mA = \mB^{\top}\mB$ after each Adam update step via a balanced projection operation 
and theoretically demonstrate that balanced low-rank factors yield fast convergence. 
However, their analysis does not take into account the effect of momentum. 
A concurrent work~\citep{ghosh2026polora} extends prior preconditioning approaches 
by bounding the per-sample loss change. 
These optimizers compute individual updates for each low-rank factor. 
The most relevant work to ours is 
Riemannion~\citep{bogachev2026lora}, 
which projects the Muon update onto the tangent space of the fixed-rank manifold. 
While our approach also adopts the fixed-rank manifold perspective, 
we develop our optimizer through the rotated optimization framework, 
with rotation and normalization as its core components, 
distinguishing it from Riemannion~\citep{bogachev2026lora}.

\pagebreak
\section{Effect of LoRA rank}
\label{sec:exp_appendix}

\begin{figure}[h]
\centering
\includegraphics[width=0.804\textwidth]{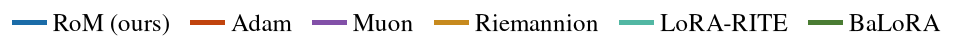}\\[2pt]
\begin{subfigure}[b]{0.45\textwidth}%
    \includegraphics[width=\linewidth]{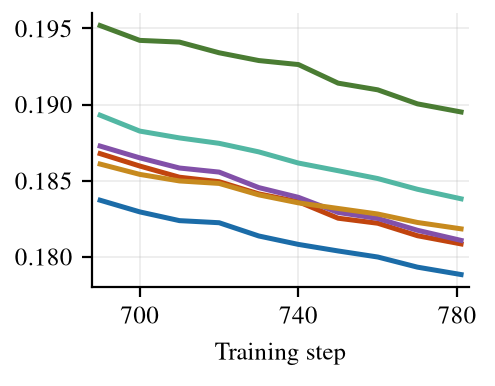}
    \caption{Llama-3.2-3B}
    \label{fig:valcurve_r32_llama3b}
\end{subfigure}%
\hspace{0.05\textwidth}%
\begin{subfigure}[b]{0.45\textwidth}%
    \includegraphics[width=\linewidth]{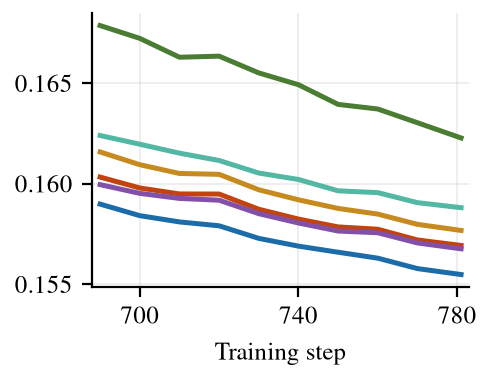}
    \caption{Llama-3.1-8B}
    \label{fig:valcurve_r32_llama8b}
\end{subfigure}%
\caption{Validation loss (mean cross-entropy across response tokens) over the last 100 training steps on the math SFT task at rank $r=32$, the counterpart of Figure~\ref{fig:valcurve_r16}.}
\label{fig:valcurve_r32}
\end{figure}

\begin{figure}[h]
    \centering
    \includegraphics[width=0.5\linewidth]{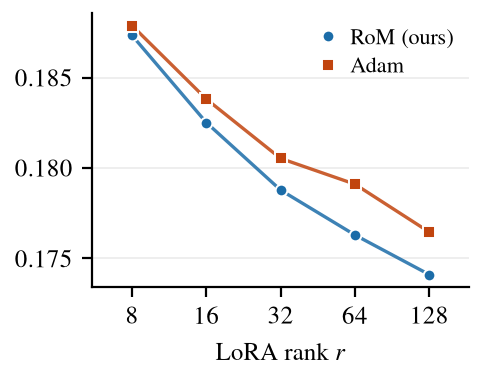}
    \caption{Final validation loss for \model and Adam across LoRA ranks on math SFT, using Llama-3.2-3B. Learning rate is independently tuned for each optimizer and rank.}
    \label{fig:rank_effect}
\end{figure}

\end{document}